\documentclass[10pt,twocolumn,letterpaper]{article}

\usepackage{cvpr}
\usepackage{times}
\usepackage{epsfig}
\usepackage{graphicx}
\usepackage{amsmath}
\usepackage{amssymb}

\usepackage[breaklinks=true,bookmarks=false]{hyperref}

\cvprfinalcopy 

\def\cvprPaperID{****} 

\begin{document}

\title{Feature Pyramid Network for Multi-Class Land Segmentation}

\author{Selim S. Seferbekov\\
Veeva Systems\\
Frankfurt am Main, 60314, Germany\\
{\tt\small selim.seferbekov@veeva.com}
\and
Vladimir I. Iglovikov\\
Level5 Engineering Center, Lyft Inc\\
Palo Alto, CA 94304, USA\\
{\tt\small iglovikov@gmail.com}
\and
Alexander V. Buslaev\\
Mapbox R$\&$D Center\\
Minsk, 220030, Belarus\\
{\tt\small aleksandr.buslaev@mapbox.com}
\and
Alexey A. Shvets\\
Massachusetts Institute of Technology\\
Cambridge, MA 02139, USA\\
{\tt\small shvets@mit.edu}
}

\maketitle

\begin{abstract}
   Semantic segmentation is in-demand in satellite imagery processing. Because of the complex environment, automatic categorization and segmentation of land cover is a challenging problem. Solving it can help to overcome many obstacles in urban planning, environmental engineering or natural landscape monitoring. In this paper, we propose an approach for automatic multi-class land segmentation based on a fully convolutional neural network of feature pyramid network (FPN) family. This network is consisted of pre-trained on ImageNet Resnet50 encoder and neatly developed decoder. Based on validation results, leaderboard score and our own experience this network shows reliable results for the DEEPGLOBE - CVPR 2018 land cover classification sub-challenge. Moreover, this network moderately uses memory that allows using GTX 1080 or 1080 TI video cards to perform whole training and makes pretty fast predictions.    
\end{abstract}

\section{Introduction}
The advent of high-resolution optical satellite imagery opens new possibilities to monitor changes on the earth's surface. The advantages of this data compared to aerial imagery are the almost worldwide availability, and sometimes the imagery data contains additional spectral channels \cite{pesaresi2001new, iglovikov2017satellite}. The geometric resolution with 0.5-1.0 m is worse than for aerial imagery, but for the land cover categorization, it is sufficient \cite{spacenet_dataset, demir2018deepglobe}. The worldwide availability of the data makes it possible to produce topographic databases for nearly any region of the earth. It, in turn, can help in various industries to enhance their productivity and quality of work whether it is for military purposes of disaster prevention or relief. 

In the last years, different methods have been proposed to tackle the problem of creating convolutional neural networks (CNN) that can produce a segmentation map for an entire input image in a single forward pass. One of the most successful state-of-the-art deep learning method is based on the Fully Convolutional Networks (FCN) \cite{long2015fully}. The main idea of this approach is to use CNN as a powerful feature extractor by replacing the fully connected layers by convolution one to output spatial feature maps instead of classification scores. Those maps are further upsampled to produce dense pixel-wise output. This method allows training CNN in the end to end manner for segmentation with input images of arbitrary sizes. Moreover, this approach achieved an improvement in segmentation accuracy over common methods on standard datasets like PASCAL VOC \cite{everingham2015pascal}. 

FCN has been further improved and now known as U-Net and Feature Pyramid (FPN) neural networks \cite{ronneberger2015u, lin2017feature, kirillov2017unified}. We approach the problem of multi-class land segmentation using FPN. FPN uses a pyramidal hierarchy of deep convolutional networks to construct feature pyramids with marginal extra cost. It consisted of bottom-up and top-down pathways. For the  bottom-up feature encoder we have chosen to use ResNet50 \cite{he2016deep} pre-trained on ImageNet.  A top-down pathway with lateral connections is developed for building high-level semantic feature maps at all scales. This architecture shows significant improvement as a generic feature extractor in several applications such as object detection and instance object segmentation \cite{lin2017focal, he2017mask, shvets2018automatic, shvets2018angiodysplasia}. Specific applications of convolutional networks on remote sensing image segmentation have been recently studied \cite{iglovikov2018ternausnet, iglovikov2018ternausnetv2, liu2017hourglass, volpi2017dense, chen2018semantic}. These applications are based mostly on aerial imagery data and contain pretty simple network architectures. In this work, we generalize many ideas and provide a neural network for semantic land segmentation that can work with high-resolution satellite imagery and approach problem of multi-class segmentation.

\section{Dataset}
\begin{figure}[t]
\begin{center}
\includegraphics[width=\linewidth]{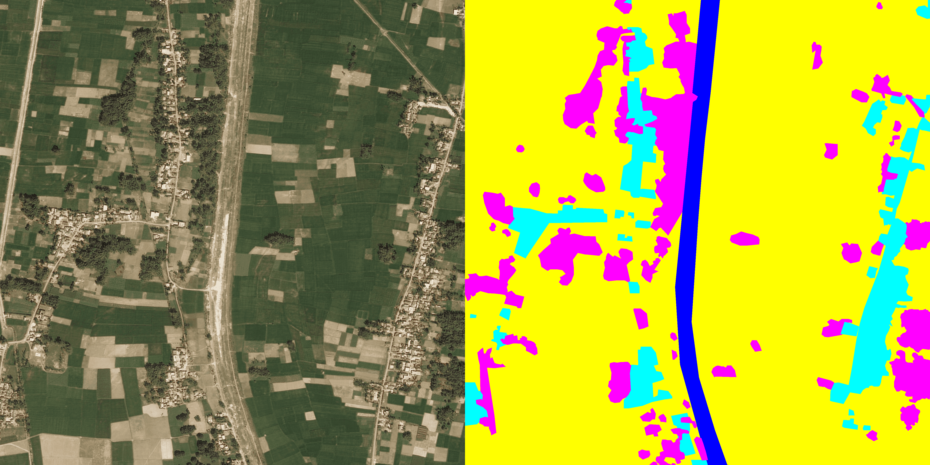}
\end{center}
   \caption{A satellite image (left) taken from the dataset and corresponding multi-channel mask (right). Each color on the mask indicate different class of objects.}
\label{fig::lands}
\end{figure}

The training data for land cover challenge contains 803 satellite imagery in RGB format. Each image has a size of 2448x2448 pixels. These images have 50cm pixel resolution, collected by DigitalGlobe's satellite \cite{deepglobe_website, demir2018deepglobe}. Moreover, each image in the training dataset contains a paired mask for land cover annotations. The mask is given as a RGB image with 7 classes of labels, using color-coding (R, G, B) as follows: 1) Urban land: (0, 255, 255) - man-made, built up areas with human artifacts (without roads); 2) Agriculture land: (255, 255, 0) - farms, any plantation, cropland, orchards, vineyards, nurseries, and ornamental horticultural areas; confined feeding operations; 3) Rangeland: (255, 0, 255) - any non-forest, non-farm, green land, grass; 4) Forest land: (0, 255, 0) - any land with certain tree crown density plus clear-cuts; 5) Water: (0, 0, 255) - rivers, oceans, lakes, wetland, ponds; 6)  Barren land: (255, 255, 255) - mountain, land, rock, dessert, beach, no vegetation; 7) Unknown: (0, 0, 0) - clouds and others. A satellite image and corresponding mask from the training dataset is shown in Fig.\ref{fig::lands}.

It is worth to mention that the values of the mask image may not be pure 0 and 255. As a result, the recommended threshold for binarization is 128 for each label. Moreover, the labels are not perfect due to the high cost of annotating multi-class segmentation mask. In addition, small objects are not annotated consciously. To measure the performance of our model we also provided by 171 validation images that do not contain masks. The predicted masks for the validation images should be upload to deepglobe website \cite{deepglobe_website}.

\section{Model}
\begin{figure*}
\begin{center}
\includegraphics[width=\textwidth,height=9cm]{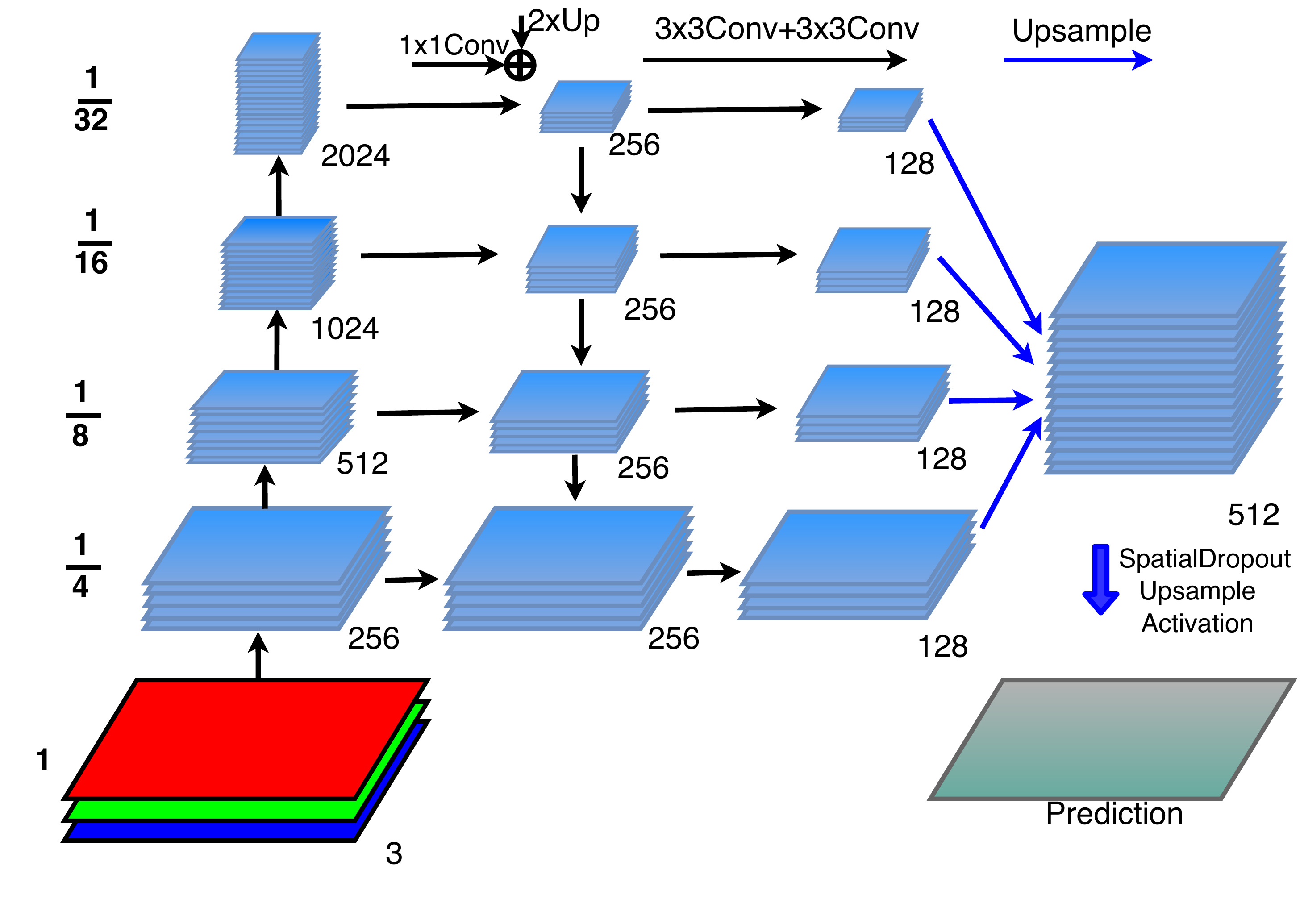}
\end{center}
   \caption{Feature pyramid network with Resnet50 encoder pre-trained on ImageNet. As an input, we have an RGB image. The number of channels increases stage by stage on the left part of the scheme while the size of the feature maps decreases stage by stage. The arrows on top show transformations implemented between the layers. In the final step, feature maps upsample to the same size and concatenated. Then, the number of channels decreases to the number of classes, and the resulting image is upsampled to the original image size.}
\label{fig::fpn}
\end{figure*}

Objects segmentation in different scales is challenging in particular for small objects. For this problem, we use feature pyramid network (FPN) to implement land segmentation. The general scheme for FPN is shown in Fig. \ref{fig::fpn}. FPN composes of a bottom-up and top-down pathways. The bottom-up pathway is the typical convolutional network for feature extraction. In our particular case, we have chosen to use pre-trained ResNet50 as a feature encoder. It composes of many convolution modules ($conv_i$ for $i$ equals 1 to 5) each has many convolution layers. As we move up, the spatial dimension is reduced by 1/2 (i.e., double the stride). The output of each convolution module is labeled as $C_i$ and later used in the top-down pathway. 

As we go up, the spatial resolution decreases. With more high-level structures extracted, the semantic value for each layer increases. On the top of the bottom-up pathway, we apply a 1x1 convolution filter to reduce $C_5$ channel depth to 256 to create $M_5$ module. Then we apply successively two times 3x3 convolution to create $P_5$ which becomes the first feature map layer used for object segmentation. $P_5$ has the same spatial resolution as $conv_5$ and 128 channels. As we go down in the top-down path, we upsample the previous layer by 2 using nearest neighbors upsampling. We again apply a 1x1 convolution to the corresponding feature maps in the bottom-up pathway. Then we add them element-wise. After that, we apply successively two times 3x3 convolution to output the next feature map layers for object segmentation. This filter reduces the aliasing effect of upsampling.

In the final step, we concatenate all $P_i$ modules that have 1/4 of the input image resolution and 128 channels each. As a result, we have a module with 512 channels. Then, we apply 512 3x3 convolution filters, batch normalization, and ReLU activation. To reduce over-fitting we apply spatial dropout operation \cite{tompson2015efficient}. As a next step, we reduce the number of output channels utilizing 1x1 convolutions. In this problem, we have seven classes so that the output feature map has seven channels. Then up-sample output to the original image size using bi-linear interpolation.   

\section{Training}
As the first step in training, we prepared masks as seven channel images using one hot encoding. Then,  as the evaluation metric, we use Jaccard index (Intersection Over Union or IoU). It can be interpreted as a similarity measure between a finite number of sets. For two sets $A$ and $B$, it can be defined as following:
\begin{equation}
\label{jaccard_iou}
    J(A, B) = \frac{|A\cap B|}{|A\cup B|} = \frac{|A\cap B|}{|A|+|B|-|A\cap B|}
\end{equation}
Since an image consists of pixels, the last expression can be adapted for discrete objects in the following way:
\begin{equation}
\label{dicrjacc}
J=\frac{1}{n}\sum_{c=1}^7w_c\sum\limits_{i=1}^n\left(\frac{y_i^c\hat{y}^c_i}{y_{i}^c+\hat{y}^c_i-y_i^c\hat{y}_i^c}\right)
\end{equation}
where $y_i^c$ and $\hat{y}_i^c$ are a binary values (label) and corresponding predicted probability for the pixel $i$ of the class $c$. In addition, we introduce the class weights $w_c$ that could help us to prevent difficulties with classes imbalance. For simplicity, in this problem we set $w_c = 1$ for $c\in[1, \dots 7]$. 

Since image segmentation task can also be considered as a pixel classification problem, we additionally use common classification loss functions, denoted as $H$ that for a multi-class segmentation problem is a categorical cross entropy.

The final expression for the generalized loss function is obtained by combining Eq. (\ref{dicrjacc}) and $H$ as following:
\begin{equation}
\label{free_en}
L=\alpha H+\beta(1-J)
\end{equation}
By minimizing this loss function, we simultaneously maximize probabilities for right pixels to be predicted and maximize the intersection $J$ between masks and corresponding predictions. For the land classification problem, using validation technique, we found $\alpha = 1$ and $\beta = 0.5$ that provided the best performance. 

For training our network, we split our dataset using 1/4 hold out values for validation. Then, on the fly, we make several augmentations to increase the diversity of our data artificially. For spatial augmentation, we use scale transformations $0.6-1.4$ of the original image and mask. Then, we randomly rotate the image and mask by 30 degrees. From the resulting image and mask, we take random crops with size 448x448 pixels. These images are subject to color transformation such as random contrast/brightness/HSV. One video card GTX1080$Ti$ with 11 Gb of memory allows using the batch size of 8 images.  

We train our network using Adam optimizer with learning rate 1e-4 and decay 1e-4 \cite{kingma2014adam}. The training is done for 20k iterations (batches) saving weights from several best iterations. Since the dataset is fairly limited in its size and labels of train images are not robust the predicted value for IoU is varied significantly between iterations as well as between classes. To reduce the effect of over-fitting, we use spatial dropout operation on the output of our network with $p=0.5$ \cite{tompson2015efficient}. After that, the training and validation IoU became close to each other at value $0.55$ with a standard deviation around $0.13$ between classes and standard deviation $0.03$ between iterations. 

We made predictions on the whole image with 2448x2448 pixels padding it by 8 pixels so that the side is divisible by $32=2^5$. It helped to prevent artifacts related to the bottom-up pathway of the network. To improve the robustness of our predictions, we also implemented test time augmentation (TTA) that composed of averaging of 4 predictions that correspond to 90 degrees rotation each. 

\section{Conclusions}
We developed multi-class land segmentation algorithms using feature pyramid network with ResNet50 network pre-trained on Imagenet in the bottom-up pathway and a neatly designed loss function. The main difficulty in this multi-class problem come from inaccurate labeling of classes in the training dataset. To prevent over-fitting, we used pretty strong spatial dropout on the last layer of the network as well as test time augmentation technique. The best public score of our model on the public leaderboard is 0.493. 

\section*{Acknowledgment}
The authors would like to thank Open Data Science community \cite{ods_website} for many valuable discussions and educational help in the growing field of machine/deep learning.

{\small
\bibliographystyle{ieee}
\bibliography{landlib}
}

\end{document}